# Role of Morphology Injection in Statistical Machine Translation

SREELEKHA. S, Indian Institute of Technology Bombay, India
PUSHPAK BHATTACHARYYA, Indian Institute of Technology Bombay, India

Phrase-based Statistical models are more commonly used as they perform optimally in terms of both, translation quality and complexity of the system. Hindi and in general all Indian languages are morphologically richer than English. Hence, even though Phrase-based systems perform very well for the less divergent language pairs, for English to Indian language translation, we need more linguistic information (such as morphology, parse tree, parts of speech tags, etc.) on the source side. Factored models seem to be useful in this case, as Factored models consider word as a vector of factors. These factors can contain any information about the surface word and use it while translating. Hence, the objective of this work is to handle morphological inflections in Hindi and Marathi using Factored translation models while translating from English. SMT approaches face the problem of data sparsity while translating into a morphologically rich language. It is very unlikely for a parallel corpus to contain all morphological forms of words. We propose a solution to generate these unseen morphological forms and inject them into original training corpora. In this paper, we study factored models and the problem of sparseness in context of translation to morphologically rich languages. We propose a simple and effective solution which is based on enriching the input with various morphological forms of words. We observe that morphology injection improves the quality of translation in terms of both adequacy and fluency. We verify this with the experiments on two morphologically rich languages: Hindi and Marathi, while translating from English.

Morphology Injection; a case study on Indian Language perspective

· Computing methodologies → Artificial intelligence → Natural language processing → Machine translation    · Computing methodologies → Artificial intelligence → Natural language processing → Phonology / morphology

Additional Key Words and Phrases: Statistical Machine Translation, Factored Machine Translation Models, Morphology Injection.

## 1. INTRODUCTION

Formally, Machine translation is a subfield of computational linguistics that investigates the use of software to translate text or speech from one natural language to another[1]. Languages do not encode the same information in the same way, which makes machine translation a difficult task. The Machine Translation methods are classified as transfer-based, rule-based, example-based, interlingua-based, statistics-based, etc. Statistical machine translation (SMT) is a machine translation paradigm where translations are generated on the basis of statistical models whose parameters are derived from the analysis of bilingual text corpora[2].

---

This work is supported by Department of Science & Technology, Government of India under Woman Scientist Scheme (WOS-A) with the project code – "SR/WOS-A/ET/1075/2014".

Author's addresses: Sreelekha. S, DST- Woman Scientist, Dept. of Computer Science and Engineering, Indian Institute of Technology, Bombay, India, Email: sreelekha@cse.iitb.ac.in, Piyush Dungarwal, Indian Institute of Technology Bombay, India, Email: piyushdd@cse.iitb.ac.in, Pushpak Bhattacharyya, Vijay & Sita Vashee Chair Professor, Indian Institute of Technology, Bombay, India, Email: pb@cse.iitb.ac.in.

[1] http://en.wikipedia.org/wiki/Machine_translation
[2] http://en.wikipedia.org/wiki/Statistical_machine_translation

- Word-based models: The basic unit of translation is a word. IBM models 1 to 5 describe these models. Even though these models are simple, their biggest disadvantage is that they do not consider context while modeling.

- Phrase-based models: The aim is to reduce the restrictions of word-based models by translating chunks of words which are contiguous, also called Phrases. Note that these phrases need not be linguistic phrases. The length of the phrase is variable. Phrase-based models are currently most widely used models for the SMT.

- Syntax-based models: Syntax-based translation is based on the idea of translating syntactic units, rather than single words or strings of words as in phrase- based MT. These models make use of syntactic features of a sentence such as parse trees, parts of speech (POS) tags, etc.

- Hierarchical phrase-based models: Hierarchical phrase-based translation combines the strengths of phrase-based and syntax-based translation. It uses phrases (segments or blocks of words) as units for translation and uses synchronous context-free grammars as rules (syntax-based translation).

- Factored phrase-based models: Phrase-based models are a special case of factored models. Factored models make use of vector of factors which may rep- resent morphological or syntactic information about that phrase instead of just using surface form of phrase. Even though Factored models try to add-in linguistic support for statistical approach, data sparseness and increased decoding complexity are big road-blocks in their development.

Statistical translation models which translate into a morphologically rich language face two challenging tasks:

- **Correct choice of inflection:** As single source root word can be mapped to several inflectional forms of target root word, the translation system should get the missing information from the source text that can help to make correct inflectional choice

- **Data sparsity:** During training, the corpus of morphologically rich language cannot have all inflectional forms of each word

Most approaches to Statistical Machine Translation, i.e., phrase based models (Koehn, Och and Marcu, 2003), syntax based models (Yamada and Knight 2001) do not allow incorporation of any linguistic information in the translation process. The introduction of factored models (Koehn and Hoang, 2007) provided this missing linguistic touch to the statistical machine translation. Factored models (Koehn and Hoang, 2007) treat each word in the corpus as vector of tokens. Each token can be any linguistic information about the word which leads to its inflection on the target side. Hence, factored models are preferred over phrase based models (Koehn, Och and Marcu, 2003) while translating from morphologically poor language to morphologically richer language.

Factored models translate using *Translation* and *Generation* mapping steps. If a particular factor combination in these mapping steps has no evidence in the training corpus, then it leads to the problem of data sparseness. Hence, though factored models give more accurate morphological translations, they may also generate more unknowns compared to other unfactored models. In this paper, we study factored models and the problem of sparseness in the context of translation to morphologically rich languages. We propose a simple and effective solution which is based on enriching the input with various morphological forms of words.

To understand the severity of the sparseness problem, we consider an example of verb morphology in Hindi[3]. Hindi verbs are inflected based on gender, number, person, tense, aspect and modality. Gender has two values (masculine, non-masculine). Number has two values (singular and plural). Person has three values (first, second and third). Tense has two values (present, non-present). Aspect has three values (simple, progressive and perfect). Modality has around nine values (shall, will, can, etc.). Thus, for a single root verb in Hindi, we have in total 648 (2*2*3*2*3*9) inflected forms of it. Hence, a single English verb can be translated to 648 verbs in Hindi side. Hindi vocabulary has around 40,000 root verbs. Hence, in total 25,920,000 (648*40,000) verb forms. It is very likely that parallel Hindi corpus cannot have all inflected forms of each verb. Also note that, if the corpus size of morphologically richer language is lesser then the problem of sparseness will be more severe.

Thus, even though we can use factored models to correctly generate morphological forms of words, the problem of data sparseness limits their performance. In this paper, we propose a simple and effective solution which is based on enriching the input corpora with various morphological forms of words. Application of the technique to English-Hindi case-study shows that the technique really improves the translation quality and handles the problem of sparseness effectively.

The rest of the paper is organized as follows: Section 2 describes the related work; Section 3 describes the overview of Hindi Inflectional Morphology; Section 4 describes the basics of factored translation models; Section 5 discusses the data sparseness problem and the proposed solution; Section 6 discusses Morphology Injection technique; Section 7 discusses Resource Generation process; Section 8 discusses Morphology Generation process; Section 9 discusses Experiments and evaluations conducted; Section 10 gives a generalized solution to the sparseness problem; Section 11 draws conclusion and points to future work.

## 2. RELATED WORK

Since India is rich in linguistic divergence there are many morphologically rich languages quite different from English as well as from each other, there is a large requirement for machine translation between them. Development of efficient machine translation systems using appropriate methodologies and with limited resources is a challenging task. There are many ongoing attempts to develop MT systems for Indian languages (Antony, 2013; Kunchukuttan et al., 2014; Sreelekha et al., 2014; Sreelekha et al., 2015) using both rule based and statistical approaches. There were many attempts to improve the quality of Statistical MT systems such as;

[3] Hindi and Marathi are morphologically rich languages compared to English. They are widely spoken in Indian sub- continent.

using Monolingually-Derived Paraphrases (Marton et al., 2009), using Related Resource-Rich languages (Nakov and Ng, 2012). Considering the large amount of human effort and linguistic knowledge required for developing rule based systems, statistical MT systems became a better choice in terms of efficiency. Still the statistical systems fail to handle rich morphology.

Research on handling rich morphology has largely focused on translating from rich morphology languages, such as Arabic, into English (Habash and Sadat, 2006). There has been less work on translating from English into morphologically richer languages. Koehn (2005) reports in a study of translation quality for languages in the Europarl corpus, that translating into morphologically richer languages is more difficult than translating from morphologically richer languages. There are valid reasons why generating richer morphology from morphologically poor languages is harder. Consider the example of translating noun phrases from English to Hindi (or German, Czech, etc.). In the case of English, a noun phrase is rendered the same if it is the subject or the object. On the other hand, noun phrases are inflected based on their role in the sentence in Hindi words. A purely lexical mapping of English noun phrases to Hindi noun phrases suffers from the lack of information about its role in the sentence, making it hard to choose the right inflected forms.

In one of the first efforts to enrich the source in word-based SMT, Ueffing and Ney (2003) used part-of-speech (POS) tags, in order to deal with the verb conjugation of Spanish and Catalan; so, POS tags were used to identify the pronoun+verb sequence and splice these two words into one term. The approach was clearly motivated by the problems occurring by a single-word-based SMT and has been solved by adopting a phrase-based model. Meanwhile, there is no handling of the case when the pronoun stays in distance with the related verb. Minkov et al. (2007) suggested a post-processing system which uses morphological and syntactic features, in order to ensure grammatical agreement on the output. The method, using various grammatical source-side features, achieved higher accuracy when applied directly to the reference translations but it was not tested as a part of an MT system. Similarly, translating English into Turkish (Durgar El-Kahlout and Oflazer, 2006) uses POS and morph stems in the input along with rich Turkish morph tags on the target side, but improvement was gained only after augmenting the generation process with morphotactical knowledge. Habash et al. (2007) also investigated case determination in Arabic. Carpuat and Wu (2007) approached the issue as a Word Sense Disambiguation problem.

Koehn and Hoang (2007) have conducted experiments on factored SMT models using morphology tags added on the morphologically rich side and scored with a 7-gram sequence model, along with POS tags for translating from English to German, Spanish and Czech. Birch et al. (2007) investigated the probabilistic models for using only source tags, where English was the target language. They have used Combinatorial Categorial Grammar (CCG) supertags as factors on the input words in factored SMT models.

Although past work focuses on studying complexity of factored translation models ( Tamchyna and Bojar, 2013), the problem of data sparseness is not addressed, to the best of our knowledge. Also, substantial volume of work has been done in the field of translation into morphologically rich languages. The source language can be enriched with grammatical features (Avramidis and Koehn, 2008) or standard translation model can be appended with *synthetic phrases* (Chahuneau et al., 2013). We discuss a case study in which we try to handle the noun morphology in English to Hindi translation using factored models. There has been previous work done in order to solve the verb morphology for English to Hindi translation (Gandhe et al., 2011). The goal is to handle data sparseness against this case study. Our experiments show that the model performs very well in order to handle the noun morphology and solving the sparseness problem.

### 3. OVERVIEW OF HINDI INFLECTIONAL MORPHOLOGY

Hindi is morphologically rich language compared to English. Hence, for building better English-Hindi translation system, we need to know how Hindi morphology works. In this Section, we get a brief overview of the Hindi noun and verb-based inflectional morphology.

#### 3.1 Verb morphology

| Verbal Form | Grammatical Category | | Exponents |
|---|---|---|---|
| Finite | Tense | Present | ho |
| | | Past | th- |
| | | Future | -g- |
| | Aspect | Habitual | -t- |
| | | Progressive | rəh |
| | | Perfective | -(y)ā, -(y)ī, -(y)ĩ, -(y)e |
| | | Completive | cuk |
| | Mood | Imperative | null, -o, -iye, -jiye, -nā |
| | | Subjunctive (root) | -ũ, -o, -(y)e, -(y)ẽ |
| | | Subjunctive (auxiliary) | ho |
| | | Presumptive | ho-g- |
| | | Root conditional | -t- |
| | | Condition (auxiliary) | ho-t- |
| | Gender-Number | Masculine-singular | -(y)ā |
| | | Masculine-plural | -(y)e |
| | | Feminine-singular | -(y)ī |
| | | Feminine-plural | -(y)ī |
| | Person-Number | 1st p-singular | -ũ |
| | | 1st p-plural | -(y)ẽ |
| | | 2nd p-singular | -(y)e<br>-o (semi-honorific) |
| | | 2nd p-plural | -o (semi-hon)<br>-(y)ẽ (honorific) |
| | | 3rd p-singular | -(y)e |
| | | 3rd p-plural | -(y)ẽ |
| | Voice | Passive | Perfective + auxiliary 'jā' |
| Non-Finite | Infinitive | | -n- |
| | Past Participle | | -(y)ā, -(y)ī,-(y)ĩ,-(y)e |
| | Present Participle | | -t- |

Figure 1: Inflectional categories and their markers for Hindi verbs [Singh and Sarma, 2011]

Many grammarians and morphologists have discussed the inflectional categories of Hindi verbs but these studies are either pedagogical or structural in approach. Across all approaches, there is much agreement on the kinds of inflectional categories that are seen in Hindi verbs. The inflection in Hindi verbs may appear as suffixes or as auxiliaries [Singh and Sarma, 2011]. These categories and their exponents are shown in Figure 1.

While translating from English to Hindi to handle all of these inflections of verbs we need to have all the factors available with us to implement a factored model. But, as we will see, so many factors in factored model may degrade the performance of the translation system. Hence, we tried to use some of these factors which are important and which are easily available.

### 3.2 Noun morphology

Hindi nouns show morphological marking only for number and case. Number can be either singular or plural. Case marking on Hindi nouns is either direct or oblique. Gender, an inherent, lexical property of Hindi nouns (masculine or feminine) is not morphologically marked, but is realized via agreement with adjectives and verbs [Singh and Sarma, 2010]. Morphological classification of the noun is shown in Figure 2. All nouns in the same class have same morphological inflection behavior. Nouns are classified into five classes [Singh and Sarma, 2010].

|          | Class A | Class B | Class C | Class D | Class E |
|----------|---------|---------|---------|---------|---------|
| **Sg-dir** | *null* | *null* | *null* | *null* | *null* |
| **Sg-obl** | *null* | *null* | *null* | *-e* | *null* |
| **Pl-dir** | *null* | *-yā̃* | *-ē̃* | *-e* | *null* |
| **Pl-obl** | *null* | *-yõ* | *-õ* | *-õ* | *-yõ/-õ* |
| **Examples** | भूख, क्रोध, प्यार | लड़की, शक्ति, नदी | रात, माला, बहू | लड़का, धागा, भांजा | आलू, साधू, माली |

Figure 2 .  Morphological classification of Hindi nouns [Singh and Sarma, 2010]

- Class A: Includes nouns that take null for all number-case values.  These nouns are generally abstract or uncountable.

- Class B: Includes /ī/, /i/ or /yā̄/ ending feminine nouns that take −yā̄ for the features [+pl, -dir, -oblique] and −yō for [+pl, +oblique].

- Class C: Includes feminine nouns that take −ē for the feature [+pl, dir] and −ō for [+pl, +oblique].

- Class D: Includes masculine nouns that end in /a/ or /yā/. Some kinship terms are also involved. Words directly derived from Sanskrit are excluded.

- Class E: Includes masculine nouns that inflect only for the features [+pl, + oblique]. The nouns in this class end with /ū/, /u/, /ī/, /i/ or a consonant.

*3.2.1        Predicting Inflectional Class for New Lexemes*

For the classification of the new lexemes into one of the five classes discussed in section 3.2, we need gender information. After gender is lexically assigned to the new lexeme, its inflectional class can be predicted using the procedure outlined in Figure 3. A masculine noun may or may not be inflected based on its semantic property. If it is an abstract noun or a mass noun it will fall into the non-inflecting Class A irrespective of its phonological form. On the other hand, a countable lexeme will fall into one of the two masculine classes based on its phonological form. Similar procedure follows for feminine nouns

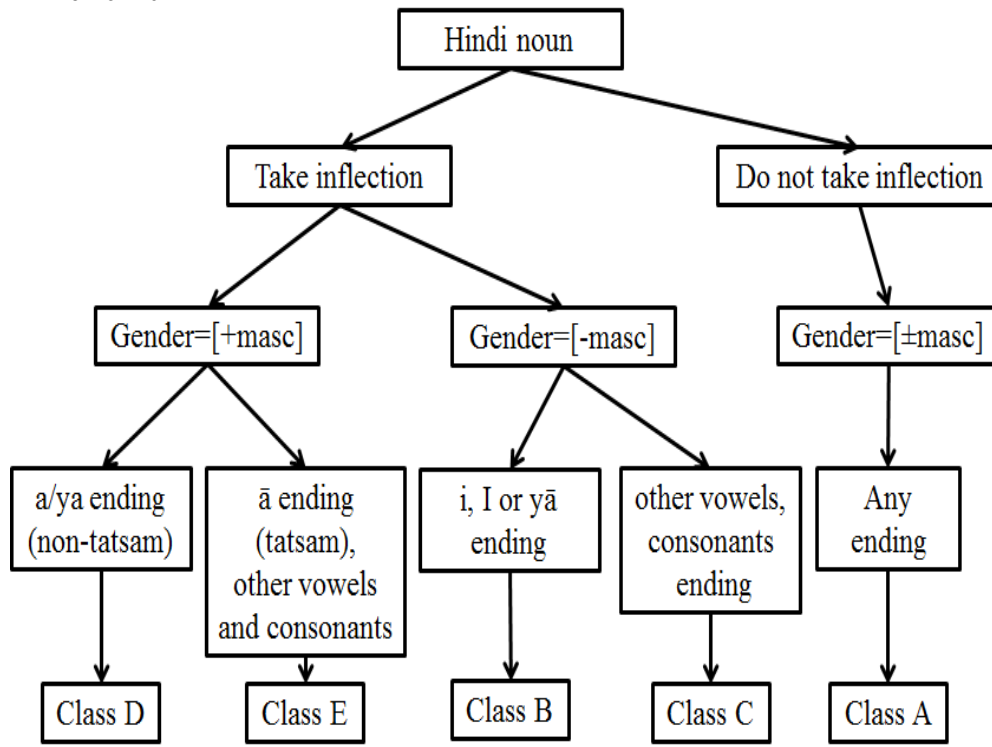

Figure 3: Predicting inflectional class for new lexemes [Singh and Sarma, 2010]

## 4.  FACTORED TRANSLATION MODELS

Phrase-based translation models are limited to mapping small contiguous word chunks without using any linguistic information such as morphology, syntax, or semantics (Koehn, Och and Marcu, 2003). On the other hand, factored translation models allow additional annotation at the word level. Factored models consider a word as a vector of tokens instead of just a single token which represents different levels of annotation as shown in Figure 4. Factored translation models can be seen as the combination of several components (language model, reordering model, translation steps, and generation steps). These components define one or more feature functions that are combined in a log-linear model (Koehn and Hoang, 2007):

$$p(e|f) = \frac{1}{Z} \sum_{i=1}^{n} \lambda_i h_i (e, f) \tag{1}$$

Each $h_i$ is a feature function for a component of the translation, and the $\lambda_i$ values are weights for the feature functions. Z is normalization constant.

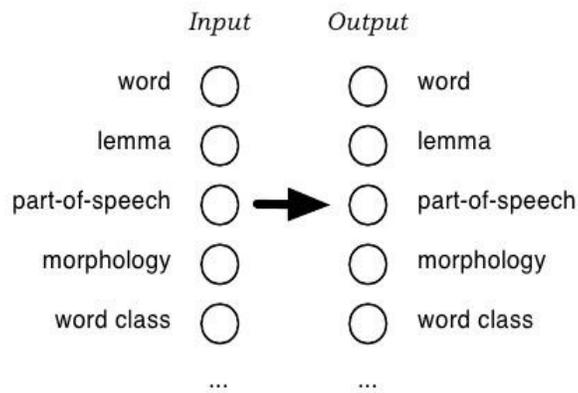

Figure 4: Factored representations of input and output words (Koehn and Hoang, 2007)

### 4.1 Factored model for handling morphology

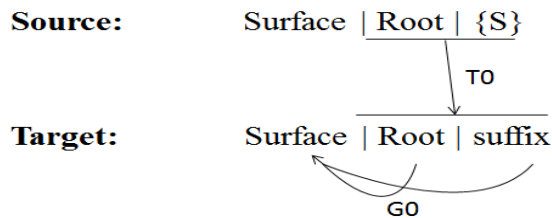

Figure 5: Factored model setup to handle nominal inflections

Note that our goal is to solve the sparseness problem while translating to morphologically rich languages. Figure 5 shows a general factored model approach for translation from morphologically poor language to morphologically rich language. On the source side we have: Surface word, root word, and set of factors $S$ that affect the inflection of the word on the target side. On the target side, we have: Surface word, root word, and suffix (can be any inflection).

The model has the following mapping steps:

- **Translation step (T0):** Maps source root word and factors in $S$ to target root word and target suffix

- **Generation step (G0):** Maps target root word and suffix to the target surface word
  Note that the words which do not take inflections have *null* as values for the factors in $S$.

### 4.2 Factored models for handling verb morphology

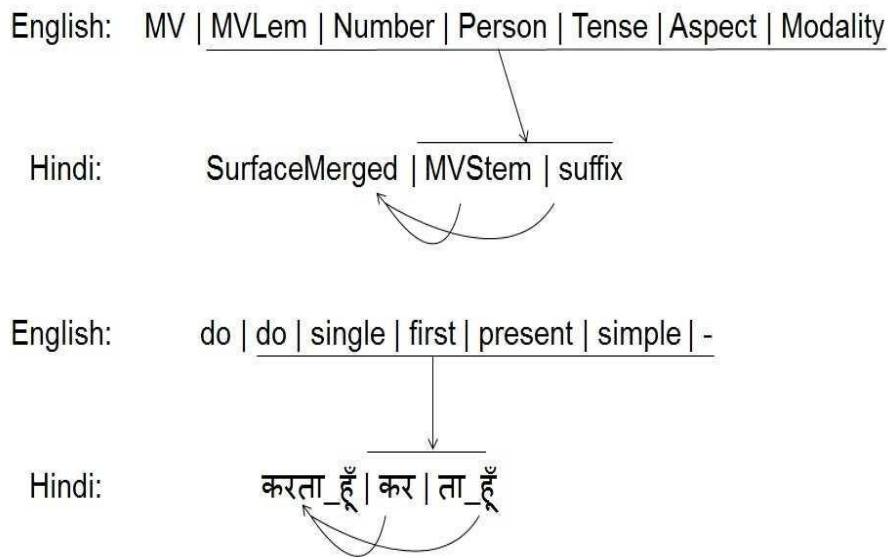

Figure 6: Factored model mapping for handling verbal inflections in Hindi

Verbal inflections in Hindi depend upon tense, number, person, gender, aspect etc. English verbs do not explicitly contain this information. Hence, while translating from English to Hindi, we need to consider syntactic and semantic information hidden in the English sentence to get this information apart from the original verb. Once we get these factors we can use factored model mapping shown in the Figure 6 to handle the morphological inflections of Hindi verbs. Gender is not used in the mapping due to two reasons. Firstly, getting gender information on English side is very hard. Secondly, just using many factors in factored model does not improve the results, but instead it may result in degradation. On English side, we use lemma of main verb only and remove any auxiliary verbs present.

Because, information contained in the auxiliaries and inflection of the verb will already be present in the other factors that we are using in factored model. For example, if a sentence has verb *is eating*, we remove *is* and retain lemma of eating, i.e., *eat*.

On Hindi side, we create a merged verb from main verb and auxiliary verbs. Main verb stem is used as a factor. We merge inflections from main verb with auxiliaries and create another factor.

Factored model has a single translation steps and single generation step:

– **Translation step:** Map main verb lemma, number, person, tense, aspect, and modality on English side to main verb stem and merged suffix on Hindi

– **Generation step:** Map main verb stem and merged suffix to surface form on Hindi side

### 4.3 Factored models for handling noun morphology

Noun inflections in Hindi broadly depend upon number and case (direct/oblique) of the noun. Hence, if we decide to use factored models for handling noun inflections, it is very natural to use number and case as factors on English side. Hence, the suggested factored model mappings would be as shown in Figure 7.

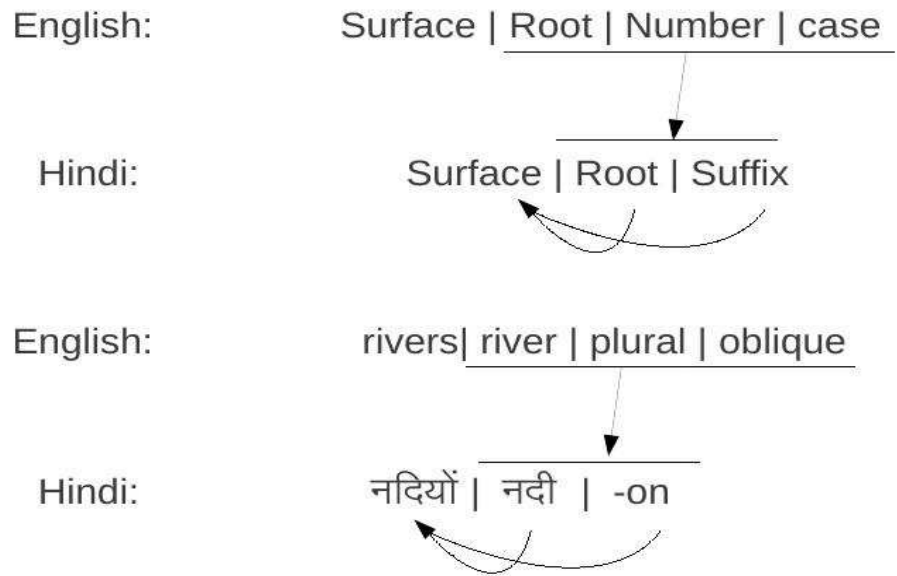

Figure 7: Factored model setup to handle nominal inflections

Factored model has a single translation step and single generation step:

- **Translation step:** Map root noun, number, and direct/oblique case on English side to root noun and suffix on Hindi side

- **Generation step:** Map root noun and suffix to surface form on Hindi side

### 4.4 Using semantic relations to generate the factors

As discussed in previous section, we need to generate tense, person, number and gender information of verb on English side. As this information is absent in the raw sentence, we need deep information about the sentence, such as parts of speech (POS) tagging, semantic relations, parse tree, etc. to generate this information. In the following subsections, we will study how to make use of these extra resources to get tense, person, number and gender information. We use Stanford dependency parser for getting syntactic parse tree of the sentence. We also use semantic relations provided by Stanford's typed dependencies (Marneffe et al. 2008). In particular, rather than the phrase structure representations that have long dominated in the computational linguistic community, typed dependencies represents all sentence relationships uniformly. That is, as triples of a relation between pairs of words, such as *the subject of distributes is Bell*.

Consider the following sentence:

> *Bell, based in Los Angeles, makes and distributes electronic, computer and building products.*

For this sentence, the Stanford Dependencies (SD) representation is:

> nsubj(makes-8, Bell-1)
> nsubj(distributes-10,  Bell-1)
> partmod(Bell-1, based-3) nn(Angeles-6, Los-5)
> prep in(based-3, Angeles-6) root(ROOT-0, makes-8)
> conj and(makes-8, distributes-10)
> amod(products-16, electronic-11)
> conj and(electronic-11, computer-13)
> amod(products-16, computer-13)
> conj and(electronic-11, building-15)
> amod(products-16, building-15)
> dobj(makes-8, products-16)
> dobj(distributes-10, products-16)

The current representation contains approximately 53 grammatical relations as described in (Marneffe et al. 2008). The dependencies are all binary relations: a grammatical relation holds between a governor (also known as a regent or a head) and a dependent.

#### 4.4.1. Tense, aspect, and modality factor

Algorithm 1 describes how to get tense, aspect, and modality of the sentence using a parse tree.

**4.4.1.1 Person Factor**

---

**Algorithm 1** Get tense, aspect, and modality of the sentence

---

**Input**: Parse tree of the sentence
**Output**: Tense, aspect, and modality of the sentence
1: tense, aspect, modality =Empty array of strings
2: For each leaf node in parse tree:
3:        POS = parent (leaf) Ψ //parts of speech (POS) tag of leaf
4:        if  (POS == "VBP"   || POS == "VBZ"   || POS == "VB")
5:        add "present" to tense
6:        else if (POS == "VBD")
7:                add "past" to tense
8:        else if (POS == "MD")
9:        if ( ! leaf == "can" && ! leaf == "could")
10:               add "future" to tense
11:      else
12:              find add auxVerb to modality
13:      else if (POS == "VBG")
14:              add "progressive" to aspect
15:      else if (POS == "VBN")
16:      add "perfect" to aspect
17: return tense, aspect, modality

---

Algorithm 2 uses typed dependency to get the subject of the sentence. Person of the subject is found by comparing subject with pronouns. If subject is not a pronoun, then most probably it will be a third person.

**4.4.1.2 Number Factor**

---

**Algorithm 2**  Get person of the sentence

---

**Input**: Parse tree of the sentence, Typed dependencies
**Output**: Person of the sentence
1: person=Empty string
2: subject = get subject using typed dependency  "nsubj"
3: if (subject in ["i", "we"])
4:        person= "first"
5: else if (subject in ["you"])
6:        person= "second"
7: else if (subject in ["he", "she", "it", "they"])
8:        person= "third"
9: else
10:        person= "third"
11: return person

---

Algorithm 3 describes how to use parts of speech (pos) tag of subject to get the number of subject. If pos tag end with s, then subject is plural, otherwise it is singular.

### *4.4.1.3      Gender factor*

---

**Algorithm 3**  Get number of the subject in the sentence

---

**Input**: Parse tree of the sentence, Typed dependencies
**Output**: Number of the subject
      number =Empty string
      subject = get subject using typed dependency "nsubj"
      POS = parent(subject)Ψ//parts of speech (POS) tag of subject
      **if** (POS.startsWith("NN") && POS.endsWith("S"))
          number= "plural"
    **else if** (POS.startsWith("NN") && ! POS.endsWith("S"))
          number= "singular"
      return number

---

Algorithm 4 describes how to get gender of the subject of the sentence. Although, this algorithm is very weak since it gets gender by comparing subject with few pronouns. Hence, other pronouns and most importantly proper nouns are not classified.

#### 4.4.1.4  Case factor

---

**Algorithm 4** Get gender of the subject of the sentence

---

**Input:** Parse tree of the sentence, Typed dependencies
**Output:** Gender of the sentence
1:  gender =Empty string
2: subject = get subject using typed dependency  "nsubj"
3: if (subject in ["he"])
4:       person= "+musc"
5: else if (subject in ["she"])
6:       person= "-musc"
7: else if (subject in ["it"])
8:       gender= "neutral"
9:  return gender

---

To get direct/oblique case of nouns on English side, we need to find out features of an English sentence that correspond to direct/oblique case of nouns in Hindi. Currently, we use following two features for this purpose.

— Object of preposition has Oblique case

E.g. Fishes live in the rivers

मछलियाँ  नदियों   में   रहती  हैं
{machhaliyan nadiyon me rahti hain}
{ fishes rivers in live}

Here, नदियों is oblique form of नदी. In the English sentence, *river* is an object of preposition *in*. Hence, we can say that object of preposition in English sentence corresponds to an oblique case of that object in parallel Hindi sentence.

— Subject of the sentence is oblique if it has a direct object and tense of the sentence is past, present perfect or past perfect

    E.g.   Boys ate apples
       लड़कों ने सेब खाए
       ladkon ne seb khaye
       boys apples ate

Here, लड़कों is oblique form of लड़का. In the English sentence, *boys* is the subject of the sentence. It has direct object, *apples*. Also, sentence has past tense.

Consider another example:

       Boys went to school
       लड़के पाठशाला गए
       ladke pathshala gaye
       boys school went

Here, लड़के is the direct form of लड़का as it is plural. (Note that, direct form of लड़का when plural and oblique form of लड़का when singular, are same, i.e., लड़के). In the English sentence, boys is the subject of the sentence. But it does not have direct object.

Algorithm 5, implements above two features to get the case of nouns.

---

Algorithm 5. Get direct/oblique case of the nouns in the sentence

---

**Input**: Parse tree of the sentence, Typed dependencies, subject, direct Object, tense
**Output**: Case of the nouns
1: case=Empty Map of strings
2: if (subject != " " && directObject != " ")
3:     if ( tense== "past" || tense== "past  perfect" || tense== "present  perfect"))
4:         Put (subject, "oblique") in case
5: For each entry dep in typed dependencies:
6:     //Object of preposition has "oblique" case
7:     if  (dep.startsWith("prep")  ||  dep.startsWith("prepc"))
8:         Put (getObject(dep), "oblique") in case
9: For all other nouns in the sentence:
10:     Put (noun, "direct") in case
11: return case

---

## 5.  PROBLEM OF DATA SPARSITY

We saw that SMT systems face the problem of data sparsity. One of the reasons is that data does not have enough inflectional forms of morphologically rich language, while translating from a morphologically poor language to a morphologically rich language. Other reason is kind of unobvious, as it arises only in the case while using factored models. In this Section, we discuss these two reasons in detail.

### 5.1.  Sparsity while translating into a morphologically rich language

Root words in morphologically rich languages have many inflectional forms. While translating from morphologically poor language to a rich language, single word in the source language can be translated to multiple words in target language. Unless training data has all such inflectional forms present, the model cannot generate correct inflectional form of the target word.

For example, consider training data has following pair of sentence:

boy plays → लड़का खेलता है  (ladaka khelta hai)

Now, for any system trained with this data, if given test input as: boy ate, the output would be: लड़का खाया (ladaka khaya). This output is wrong, as it has wrong inflection of boy.  Correct translation is: लड़के ने खाया (ladake ne khaya).

### 5.2.  Sparsity while using Factored model

While factored models allow incorporation of linguistic annotations, it also leads to the problem of data sparsity. The sparsity is introduced in two ways:

— **Sparsity in Translation**: When a particular combination of factors does not exist in the source side training corpus

For example, let the factored model have single translation step: $X|Y \rightarrow P |Q^4$. Suppose the training data has evidence for only $x_i|y_j \rightarrow p_k |q_l$ mapping. The factored model learnt from this data can not translate $x_u|y_v$, for all $u \neq i$ and $v \neq j$. The factored model generates UNKNOWN as output in these cases.

For example, suppose the training data has evidence for only SRoot1[5]|S11[6]|S21[7] → TRoot1[8]| TSuffix1[9] mapping.  The factored model (described in Section 3.4) learnt from this data can not translate SRoot1|S12|S21, SRoot1|S11|S22 or SRoot1|S12|S22. The factored model generates UNKNOWN as output.

Note that, if we train simple phrase based model on only the surface form of words, we will at least get some output, which may not be correctly inflected, but still will be able to convey the meaning.

---

[4] Capital letters indicate factors and small letters indicate values that corresponding factors can take
[5] Source root word factor: SRoot1, SRoot2, etc.
[6] Source S1 factor: S11, S12, etc.
[7] Source S2 factor: S21, S22, etc.
[8] Target root word factor: TRoot1, TRoot2, etc.
[9] Target suffix factor: TSuffix1, TSuffix2, etc.

— **Sparsity in Generation**: When a particular combination of factors does not exist in the target side training corpus

For example, let the factored model have single generation step: P | Q → R.1 Suppose the target side training data has an evidence of only $p_a | q_b \rightarrow r_c$. The factored model learnt from this data can not generate from $p_u | q_v$ all $u \neq a$ and $v \neq b$. Again the factored model generates UNKNOWN as output.

For example, suppose the target side training data has an evidence of only TSurface1[10] | TRoot1 | TSuffix1. The factored model (described in Section 3.4) learnt from this data can not generate TSurface2 from TRoot2 | TSuffix2 or TSurface3 from TRoot3 | TSuffix3.

Thus, due to sparsity, we cannot make the best use of factored models. In fact, they fare worse than the phrase-based models, especially, when a particular factor combination is absent in the training data. In the case of noun inflection factored model, this can be observed through following example:

Consider following sentence to be the training data.
  • Factored: boys | boy | plural | direct play | play | . | . → लड़के | लड़का | -e खेलते | . | . हैं | . | . (ladake khelte hain)
    Unfactored: boys play → लड़के खेलते हैं (ladake khelte hain)

Now, let the test input be: boys | boy | plural | oblique (for factored) or boys (for unfactored). As factor combination boy | plural | oblique is absent in the training data of factored model, it will generate unknown output. Whereas, even though morphologically wrong phrase-based model will generate लड़के (ladke)(boys) as output.

Hence, the use of factored models may lead to low quality translation.

### 5.3 Basic Morphology injection for solving data sparsity

The reason of data sparseness in factored models is either the combination of source side factors or target side factors are not present in the training data. So, is it possible to get all the combinations of factors in the training data? In our case, we are using three factors on source side, i.e., lemma, number and direct/oblique case and one factor on the target side, i.e., root word (Note that root word here is used for a noun with no morphological inflection. E.g.,लड़का (ladka) (boy)). And there is no generation step in our mapping; hence, sparseness due to generation step is already avoided. Now, to handle the sparseness due to translation step, we need to have all morphological forms of each root word in the training data.

Section 3.2 gives morphological classification of nouns based on number, direct/oblique case and class of the noun. Figure 2 shows the suffix that a particular noun takes based on these three factors. Hence, to generate all morphological forms of a given root word in Hindi, we need to have number, case and class of the noun to be known on English side (as we are translating from English). In the Section 4.7 we describe how to morphologically classify nouns and to generate number and case factors for nouns. Gandhe et al., [2011] try to handle verbal inflections using similar technique in which they classify the verbs into different classes. Each class has verbs which take similar inflections. After classification, they generate all the inflectional forms of verbs depending upon the class of the verb.

---

[10] Target Surface word factor: TSurface1, Tsurface2, etc.

## 6.  MORPHOLOGY INJECTION TECHNIQUE

As discussed in Section 5, to handle the sparseness of factored models, we need to generate all combinations of the factors used. In this section we will see a Morphology injection method that generates various morphological forms of noun entities by classifying them and augments the training data with newly generated morphological forms of nouns.

Basic algorithm of the Morphology injection method can be described as below:

1.  Find out noun entity pairs (Eng-Hin)
2.  Classify Hindi nouns into classes
3.  Generate new morphological forms of the nouns using the classification table
4.  Augment training data with the new forms

For example,

Let noun pair be 'river - नदी (nadi )'.  Class of Hindi noun नदी is B. Now, we generate new forms of नदी  using classification table shown in Figure 8.

river| sg | dir - नदी | नदी | Null
river| sg | obl - नदी | नदी | Null
river| pl | dir - नदियाँ  | नदी | याँ
river| pl | obl - नदियों | नदी | यों

The algorithm is elaborated in the following subsections, where it is used in two different contexts.

### 6.1. Using parallel factored corpus

We can use parallel factored corpus which has lemma, number and direct/oblique case factors on English side and root word factor on Hindi side. Generation of factors will be happening as described in Section 4.4. We need to have factored English-Hindi corpus with factors as shown in Figure 8. We pass the corpus to Noun entity identifier, which is based on the parts of speech (POS) tags to get the noun entities present in the corpus.

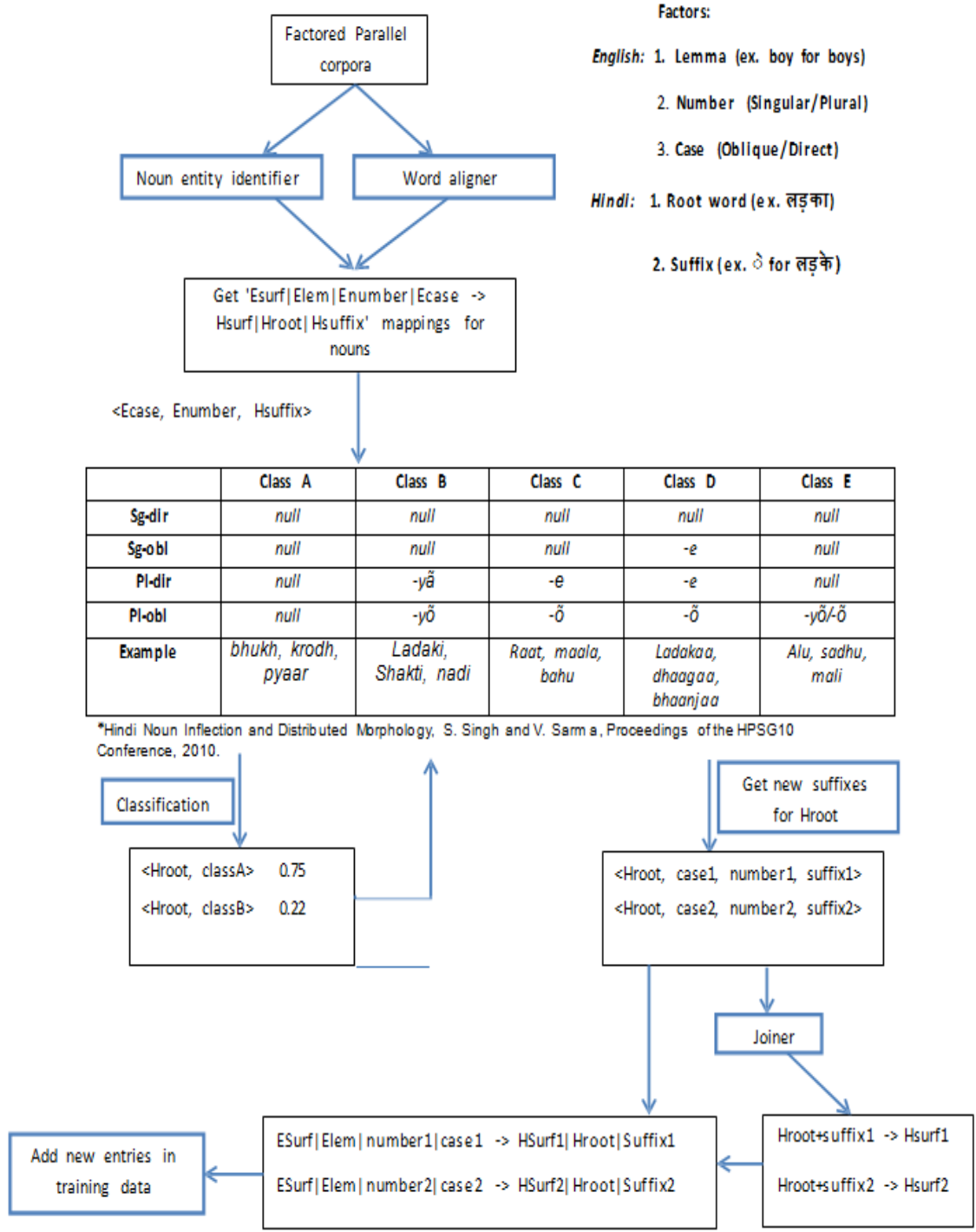

Figure 8: Using parallel factored corpus for Morphology injection method

To find out the pairs of nouns in English-Hindi corpus, we did align the corpus word by word. So, now we get the mappings of the form: Esurf|Elem|Enumber|Ecase → Hsurf|Hroot|Hsuffix for each noun pair. We can classify these noun pairs using

Enumber, Ecase and Hsuffix as discussed in Section 3.2. As each noun pair will have many corresponding combinations of number, case and suffix in the training data, we need to predict the probability of noun being classified into each of the five classes. This can be simply done by keeping a count for each class for a given noun pair and classifying each occurrence of this pair in training data into one of the classes. Note that there may be case when noun pair cannot be classified or there can be multiple classes into which the pair can be classified; in that case, we need to increment the count of each such class. Then the counts can then be normalized to get the probability. Finally, the noun pair can be classified into a class which has the highest probability.

After we classify the noun pair, we can get all combinations of number, case and suffix for given noun pair from classification table. Now, we can use these new suffixes to generate new inflected forms of the root word in Hindi. We pass new suffixes and Hindi root word to Joiner tool, which generates new surface forms. For example, given लड़का (ladka boy) and '-e' Joiner will generate लड़के (ladke)(boys). Details of the Joiner tool are discussed in Section 7.2. Finally, we get new factored pairs of the form: Esurf | Elem | Enumber | Ecase' → Hsurf | Hroot | Hsuffix'. These new pairs can be added to the original training data.

### 6.2  Using monolingual lexicon

We can also use monolingual target side lexicon to generate all combinations of factors for the factored model. In our case, we use Hindi lexicon. Hindi lexicon contains Hindi nouns, proper nouns, adjectives, verbs, etc. Figure 9 shows the pipeline of the same. The pipeline is somewhat similar to that in Figure 8, but here, instead of predicting the class of the noun pair from its suffix, we actually classify the Hindi noun into one of the five classes.

As discussed in Section 3.2, to classify a Hindi noun into a morphological class, we need its gender information, whether or not it takes inflections and its ending characters. Using this information, we can classify nouns present in the lexicon as shown in Figure 9. After classification, we can generate new morphologically inflected forms of the Hindi noun using the classification table shown in Figure 8. This process is similar to that discussed in Section 6.1. Now, we also need to generate English counterpart of the Hindi noun.

We can use Hindi-to-English dictionary for the same. After getting Englsih side root word, we can generate pairs of the form: . | Elem | Enumber' | Ecase' → Hsurf | Hroot. Note that as we cannot generate English surface word form, it is denoted by a dot in the mapping. This does not affect factored model settings, as our translation step does not use English surface word. We then append original training data with these newly generated pairs. Note that factored settings here differ from that in Section 6.1, as we do not use Hindi side suffix here.

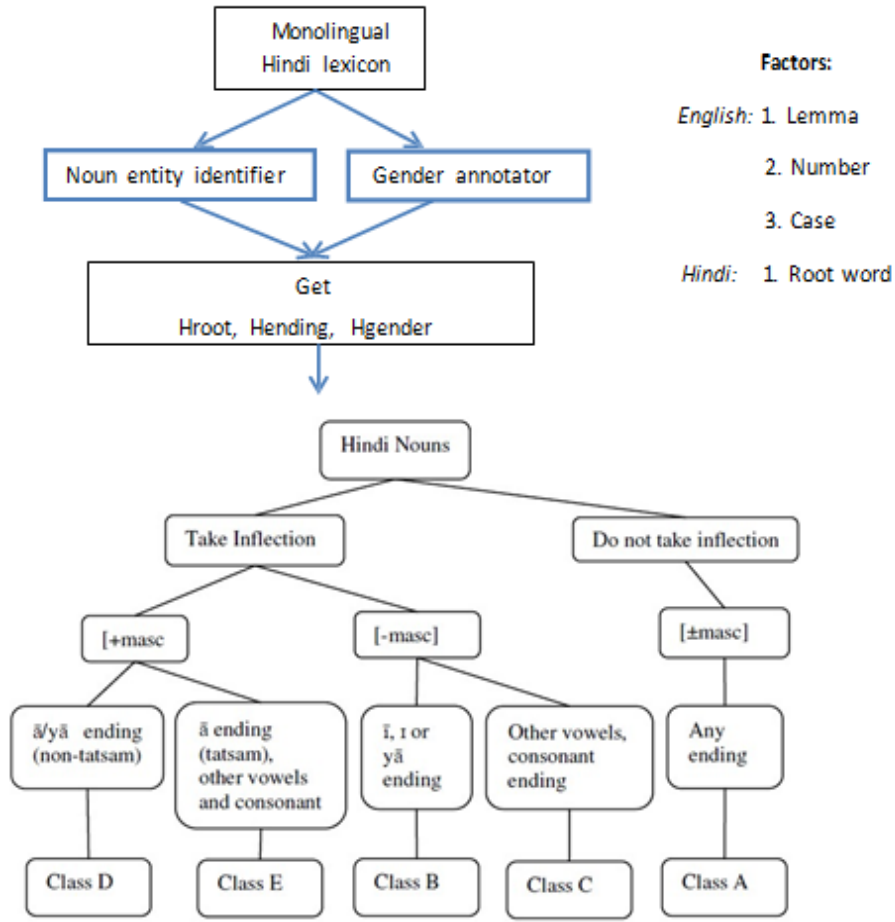

**Factors:**

*English:* 1. Lemma

2. Number

3. Case

*Hindi:* 1. Root word

*Hindi Noun Inflection and Distributed Morphology, S. Singh and V. Sarma, Proceedings of the HPSG10 Conference, 2010.

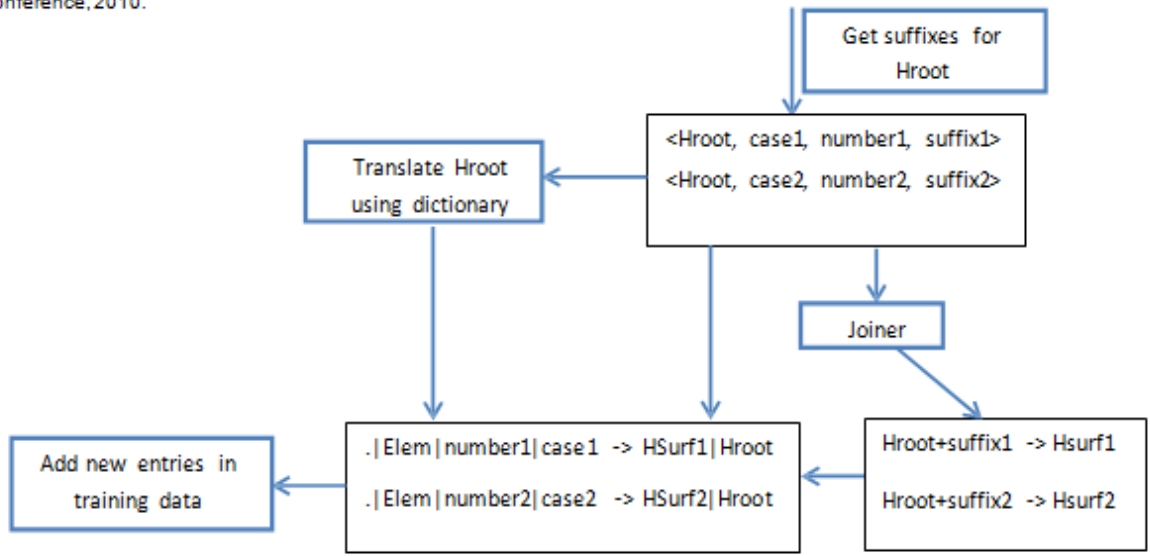

Figure 9: Using monolingual Hindi lexicon for Morphology injection method

## 7.  RESOURCE GENERATION

In this Section we discusses about the resources that need to be built, before actual training of the translation system starts.

### 7.1 Classification technique used

The approach of using parallel factored corpus as discussed in Section 6.1 is error prone and also it depends on the accuracy of the classification technique. We had Hindi lexicon readily available with us. Hence, we went forward with the approach of using Hindi lexicon for Morphology injection instead. The available Hindi lexicon size is 1,13,266 words. The lexicon has words classified into their morphological classes. Hence, we easily generated new combinations of factors, i.e., case, number and suffix for Hindi nouns as discussed in Section 6.

### 7.2 Development of a Joiner tool

After getting new suffixes for the Hindi root word, we need to form surface word by joining root word and suffix.  A rule-based joiner (or reverse morphological) tool was developed which merges root and the suffix based on the class to which the suffix belongs and the root word ending. Some of the rules are described below:

if (suffix in [यों , याँ ] (yom, yam))

     if (ending in [-e, इ , ई , ओ , ं ])

          return (root + suffix)

     else if (ending in [-ee])

          return (root - ending + -e + suffix)

For example, if the input to joiner is: नदी (nadi)(river)  and यों(yom)(s), then above rule matches for the given input. As नदी (nadi)(river) ends in -ee, output will be root - ending + -e + suffix, i.e., नदियों (nadiyon) (rivers)

Similar rules are formed for other suffixes and classes.

### 7.3 Development of a dictionary

After getting new morphological forms for Hindi root forms of the nouns, we were in need of a dictionary to translate these nouns from Hindi to English. We already had a dictionary which contained 1,28,241 Hindi-English pairs of words. But, the noun entities present in both the Hindi lexicon and the dictionary were only 9,684. Hence, instead of using this dictionary, we decided to go with an alternative approach, where we use Google's freely available online translation system to generate English nouns from Hindi. While doing this, we encountered a problem of infrequent nouns in Hindi. There were many Hindi nouns in the lexicon that were translated to same English noun. E.g. लड़का (ladka)(boy) and छोरा (chhora) (boy) are translated to boy. मछली (machhali) (fish) and मच्छी (machchhi) (fish) are translated to fish. If we use these pairs as it is, there is possibility of degrading translation as English noun may get translated to an infrequent word.

To solve the problem of infrequent words, we simply do two passes of the translation. In first pass, we translate nouns in Hindi lexicon using translation system. In the second pass, we translated these translations back to Hindi using same translation system. Hence, we get new Hindi lexicon in which the infrequent nouns are eliminated. We use these new pairs as a dictionary to translate the Hindi root words. Note that if one has frequencies of the nouns in the lexicon, they can be used directly to eliminate infrequent nouns.

## 8. MORPHOLOGY GENERATION

Hindi is a morphologically richer language compared to English. Hindi shows morphological inflections on nouns and verbs. In addition, adjectives in Hindi takes the inflection according to the gender and number of the noun it modifies. In this section, we study the problem of handling noun and verb morphology while translating from English to Hindi using factored models. We also discuss the solution to the sparseness problem.

### 8.1. Noun morphology

In this section, we discuss the factored model for handling Hindi noun morphology and the data sparseness solution in the context of same.

#### 8.1.1 Factored model setup

Noun inflections in Hindi are affected by the number and case of the noun only (Singh et al., 2010). So, in this case, the set S, as in Section 4.1, consists of number and case. Number can be singular or plural and case can be direct or oblique. Example of factors and mapping steps are shown in Figure 10. The generation of the number and case factors is discussed in Section 9.

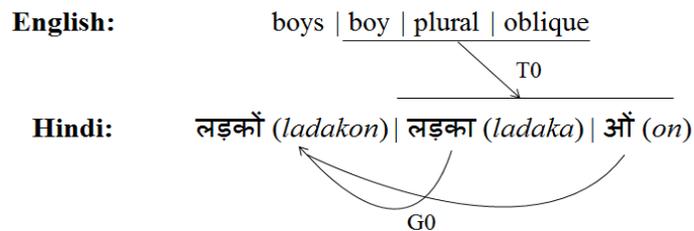

Figure 10: Factored model setup to handle nominal inflections

#### 8.1.2 Building word-form dictionary

Thus, in the case of factored model described in Section 8.1.1:

- To solve the sparseness in translation step, we need to have all English *root|number|case→Hindi noun root|number|suffix* pairs present in the training data.

- To solve the sparseness in generation step, we need to have all *Hindi noun root|number|suffix → Hindi surface word* pairs present in the training data.

In other words, we need to get a set of suffixes and their corresponding number-case values, for each noun pair. Using these suffixes and the Hindi root word, we need to generate Hindi surface words to remove sparseness in the generation step.

We need to generate four pairs for each noun present in the training data, i.e., (sg-dir, sg-obl, pl-dir, pl-obl) and get their corresponding Hindi inflections. In the following section, we discuss how to generate these morphological forms.

*8.1.2.1. Generating new morphological forms:*

Figure 11 shows a pipeline to generate new morphological forms for an English-Hindi noun pair. To generate different morphological forms, we need to know the suffix of a noun in Hindi for the corresponding number and case combination. We use the classification table shown in Table 1 for the same. Nouns are classified into five different classes, namely A, B, C, D, and E according to their inflectional behavior with respect to case and number (Singh et al., 2010). All nouns in the same class show the same inflectional behavior.

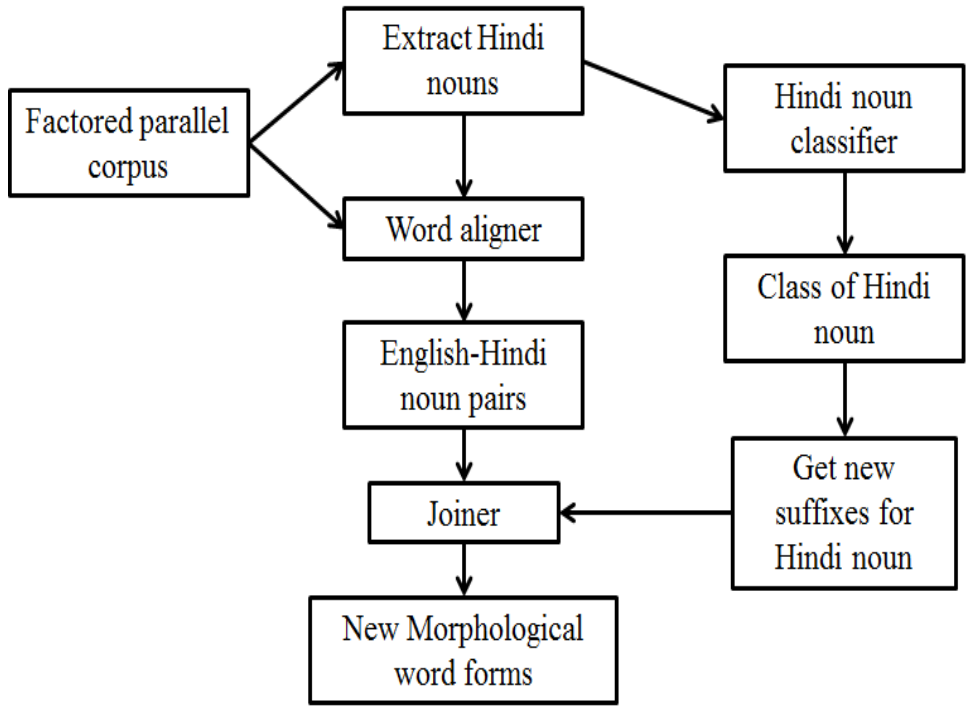

Figure 11.  Pipeline to generate new morphological forms for an English-Hindi noun pair

Table 1: Inflection-based classification of Hindi nouns (Singh et al., 2010)

|  | Class A | Class B | Class C | Class D | Class E |
|---|---|---|---|---|---|
| **Sg-dir** | *null* | *null* | *null* | *null* | *null* |
| **Sg-obl** | *null* | *null* | *null* | ए | *null* |
| **Pl-dir** | *null* | याा | एं | ए | *null* |
| **Pl-obl** | *null* | या (yam) | ओं (om) | ओं (om) | या / आ (yam /om) |

To predict the class of a Hindi noun, we develop a classifier which uses gender and the ending characters of the nouns as features (Singh et al., 2010). We get four different suffixes and corresponding number-case combinations using the class of Hindi noun and classification shown in Table 1.

For example, if we know that the noun लड़का (ladakaa)(boy) belongs to class D, then we can get four different suffixes for लड़का (ladakaa) (boy) as shown in Table 2.

*8.1.2.2. Generating surface word:*

Table 2: Morphological suffixes for boy- लड़का (ladakaa) noun pair

| **English root | Number | Case** | **Hindi root | Suffix** |
|---|---|
| boy | singular | direct | लड़का (ladakaa) | null |
| boy | singular | oblique | लड़का (ladakaa) | ए (e) |
| boy | plural | direct | लड़का (ladakaa) | ए (e) |
| boy | plural | oblique | लड़का (ladakaa) | ओं (on) |

Table 3: New morphological forms of boy- लड़का (ladakaa) noun pair

| **English root/Number/Case** | **Hindi surface/Root/Suffix** |
|---|---|
| boy/singular/direct | लड़का (ladakaa)/ लड़का (ladakaa)/null |
| boy/singular/oblique | लड़के (ladake)/ लड़का (ladakaa)/e (e) |
| boy/plural/direct | लड़के (ladake)/ लड़का (ladakaa)/e (e) |
| boy/plural/oblique | लड़कों (ladakon)/ लड़का (ladakaa)/a (on) |

Next we generate Hindi surface word from Hindi noun root and suffix using a rule-based joiner (reverse morphological) tool. The rules of the joiner use the ending of the noun root and the class to which the suffix belongs as features. Thus, we get four different morphological forms of the noun entities present in the training data. We augment the original training data with these newly generated morphological forms. Table 3 shows four morphological forms of boy- लड़का (ladakaa) noun pair. Note that the joiner solves the sparseness in generation step.

## 8.2 Verb morphology

In this section, we discuss the factored model for handling Hindi verb morphology and the data sparseness solution in the context of the same.

### 8.2.1 Factored model setup

Verb inflections in Hindi are affected by gender, number, person, tense, aspect, modality, etc. (Singh and Sarma, 2011). As it is difficult to extract gender from English verbs, we do not use it as a factor on English side. We just replicate English verbs for each gender inflection on Hindi side. Hence, set S, as in Section 4.1, consists of number, person, tense, aspect and modality. Example of factors and mapping steps are shown in Figure 12. The generation of the factors is discussed in Section 9.

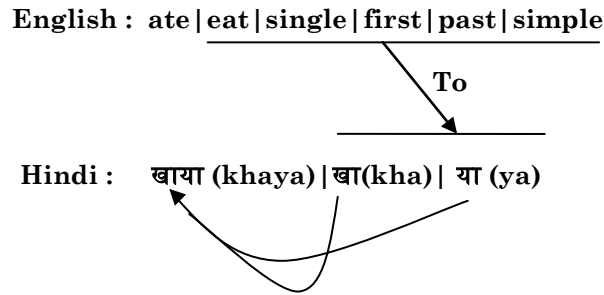

Figure 12. Factored model setup to handle verbal inflections

Here the verb ate will be having the same form खाया (khaya) in first person, second person and third person of the subject.

### 8.2.2 Building word-form dictionary

Thus, in the case of factored model described in Section 8.2.1:

- To solve the sparseness in translation step, we need to have all English *root* | *numer* | *person* | *tense* | *aspect* | *modality* →*Hindi verb root* | *suffix* pairs present in the training data.

- To solve the sparseness in generation step, we need to have all *Hindi verb root* | *suffix* → *Hindi surface word* pairs present in the training data.

In other words, we need to get a set of suffixes and their corresponding number-person-tense-aspect- modality values, for each noun pair. Using these suffixes and the Hindi root word, we need to generate Hindi surface words to remove sparseness in the generation step. In the Section 8.2.2.1, we discuss how to generate these morphological forms.

*8.2.2.1 Generating new morphological forms:*

Figure 13 shows a pipeline to generate new morphological forms for an English-Hindi noun pair. Table 4 shows a subpart of a table which is used to Figure 5: Pipeline to generate new morphological forms for an English-Hindi noun pair gets suffixes for Hindi verb roots. Note that no pre-classification of verbs is required, as these suffixes apply to all verbs. Table 5 shows few of many suffixes for भाग (bhaag).

*8.2.2.2 Generating surface word:*

Table 4: Suffixes for Hindi verbs based on number, person, tense and aspect

|         |        | Simple | |
|---------|--------|--------|--------|
|         |        | Singular | Plural |
| Present | First  | ता हूँ / ती हूँ <br> (ta hoon / ti hoon) | ते हैं / ती हैं <br> (te haen) / ti haen |
|         | Second | ता है / ती है <br> (ta hae / ti hae) | ते हो / ती हो <br> (te ho / ti ho) |
|         | Third  | ता है / ती है <br> (ta hae) / (ti hae) | ते हैं / ती हैं <br> (te haen) / (ti haen) |

Table 5: Morphological suffixes for run- भाग (bhaag) verb pair based on number (N), person (P), tense (T), aspect (A) and modality (M)

| English root | N | P | T | A | M | Hindi root | Suffix |
|---|---|
| run | singular | first | present | simple | - | भाग | ता हूँ / ती हूँ <br> (bhaag | ta hoon / ti hoon) |
| run | plural | first | present | simple | - | भाग | ते हैं / ती हैं <br> (bhaag | ta haen/ ti haen) |
| run | singular | second | present | simple | - | भाग | ता है / ती है <br> (bhaag | ta hae / ti hae) |
| run | plural | second | present | simple | - | भाग | ते हो / ती हो <br> (bhaag | te ho / ti ho) |
| run | singular | third | present | simple | - | भाग | ता है / ती है <br> (bhaag | ta hae / ti hae) |
| run | plural | third | present | simple | - | भाग | ते हैं / ती हैं <br> (bhaag | te haen / ti haen) |

Next we generate Hindi surface word from Hindi verb root and suffix using a rule-based joiner (reverse morphological) tool. The rules of the joiner use only the ending of the verb root as features. Thus, we get different morphological forms of the verb entities present in the training data. We augment the original training data with these newly generated morphological forms. Table 6 shows morphological forms of run- भाग (bhaag) verb pair. Note that the joiner solves the sparseness in generation step.

Table 6: New morphological forms of run - भाग (bhaag) verb pair

| English root/N/P/T/A/M | Hindi surface| Root / Suffix |
|---|---|
| run/singular/first/present/simple/- | भागता हूँ / ती हूँ | भाग | ता हूँ / ती हूँ <br> (bhaagta hoon/ ti hoon| bhaag | ta hoon/ ti hoon) |
| run/plural/first/present/simple/- | भागते हैं | भाग | ते हैं / <br> (bhaagte haen | bhaag | te haen) |
| run/singular/second/present/simple/- | भागता है / ती है | भाग | ता है / ती है <br> (bhaagta hae / ti hae | bhaag | ta hae/ ti hae) |
| run/plural/second/present/simple/- | भागते हो / ती हो | भाग | ते हो / ती हो <br> (bhaagte ho / ti ho | bhaag | te ho / ti ho) |
| run/singular/third/present/simple/- | भागता है | भाग | ता है / ती है <br> (bhaagta hae | bhaag | ta hae / ti hae) |
| run/plural/third/present/simple/- | भागते हैं | भाग | ते हैं / ती हैं <br> (bhaagte haen | bhaag | te haen/ ti haen) |

## 8.3 Noun and Verb morphology

Finally, we create a new factored model which combines factors on both nouns and verbs, as shown in Figure 12. We build word-form dictionaries separately as discussed in Section 8.1 and Section 8.2. Then, we augment training data with both dictionaries. Note that, factor normalization[11] on each word is required before this step to maintain same number of factors.

We also create a word-form dictionary for phrase-based model. We follow the same procedure as described in Section 8, but we remove all factors from source and target words except the surface form.

## 9. EXPERIMENTS AND EVALUATION

We performed experiments on ILCI (Indian Languages Corpora Initiative) En-Hi and En-Mr dataset. Domain of the corpus is health and tourism. We used 46,000 sentence pairs for training and 3000 sentence pairs for testing. Word-form dictionary was created using the Hindi and Marathi word lexicon. It consisted of 182,544 noun forms and 310,392 verb forms of Hindi and 44,762 noun forms and 106,570 verb forms of Marathi. Table 7 shows the statistics of the corpus used for training, testing and tuning. Table 8 shows the statistics of the generated word-form dictionary.

[11] Use *null* when particular word cannot have that factor

Table 7: Statistics of the corpus used

| Sl. No | Corpus Source | Training Corpus | Corpus Size [Parallel Sentences] |
|--------|---------------|-----------------|----------------------------------|
| 1 | ILCI | Health | 23000 |
| 2 | ILCI | Tourism | 23000 |
| Total | | | 46000 |
| Sl. No | Corpus Source | Tuning corpus(MERT) | Corpus Size [Parallel Sentences] |
| ILCI | ILCI | Tourism | 500 |
| ILCI | ILCI | Health | 500 |
| Total | | | 1000 |
| Sl. No | Corpus Source | Testing corpus | Corpus Size [Parallel Sentences] |
| 1 | ILCI | Tourism | 1500 |
| 2 | ILCI | Health | 1500 |
| Total | | | 3000 |

Table 8: Statistics of the generated word form dictionary

| Language | Verb forms generated | Noun Forms generated | Total word form dictionary size |
|----------|---------------------|---------------------|--------------------------------|
| Hindi | 310392 | 182544 | 492936 |
| Marathi | 106570 | 44762 | 151332 |

*Moses* toolkit[12] was used for training and decoding. Language model was trained on the target corpus with IRSTLM[13].

For our experiments, we compared the translation output of the following systems:

- Phrase-based (unfactored) model (Phr)

- Basic factored model for solving noun and verb morphology (Fact)

- Phrase-based model trained on the corpus used for Phr augmented with the word form dictionary for solving noun and verb morphology (Phrase -Morph)

- Factored model trained on the corpus used for Fact augmented with the word form dictionary for solving noun and verb morphology (Fact-Morph)

---

[12] http://www.statmt.org/moses/
[13] https://hlt.fbk.eu/technologies/irstlm-irst-languagemodelling-toolkit

With the help of syntactic and morphological tools, we extract the number and case of the English nouns and number, person, tense, aspect and modality of the English verbs as follows:

**Noun factors:**

- **Number factor:** We use *Stanford POS tagger*[14] to identify the English noun entities (Toutanova et al., 2003). The POS tagger itself differentiates between singular and plural nouns by using different tags.

- **Case factor:** It is difficult to find the direct/oblique case of the nouns as English nouns do not contain this information. Hence, to get the case information, we need to find out features of an English sentence that correspond to direct/oblique case of the parallel nouns in Hindi sentence. We use object of preposition, subject, direct object, tense as our features. These features are extracted using semantic relations provided by Stanfords typed dependencies (De Marneffe et al., 2008).

**Verb factors:**

- **Number factor:** Using typed dependencies we extract subject of the sentence and get number of the subject as we get it for a noun.

- **Person factor:** We do lookup into simple list of pronouns to find the person of the subject.

- **Tense, Aspect and Modality factor:** We use POS tag of verbs to extract tense, aspect and modality of the sentence.

### 9.1 Automatic evaluation

Table 9: Automatic evaluation of the translation systems for both Phrase and factor based models

| Morph Problem | Model | BLEU Score | | | | | |
|---|---|---|---|---|---|---|---|
| | | Without Tuning | | | With Tuning | | |
| | | En - Hi | En - Mr | En-Ml | En - Hi | En - Mr | En-Ml |
| Noun | Fact | 25.30 | 16.84 | 26.17 | 27.30 | 18.84 | 28.23 |
| | Fact-Morph | 31.41 | 20.85 | 32.42 | 34.41 | 22.85 | 33.45 |
| Verb | Fact | 26.03 | 17.02 | 26.54 | 28.23 | 19.52 | 28.82 |
| | Fact-Morph | 33.46 | 25.82 | 33.54 | 37.89 | 26.72 | 36.30 |
| Noun & Verb | Fact | 23.93 | 15.25 | 24.01 | 26.93 | 17.55 | 26.08 |
| | Fact-Morph | 30.03 | 23.38 | 31.56 | 33.73 | 24.58 | 32.65 |
| Noun | Phrase | 24.87 | 19.77 | 26.78 | 28.87 | 21.34 | 29.01 |
| | Phrase-Morph | 31.19 | 22.28 | 33.30 | 33.49 | 25.58 | 36.12 |
| Verb | Phrase | 25.78 | 20.17 | 26.98 | 28.87 | 22.27 | 29.17 |
| | Phrase-Morph | 32.29 | 23.28 | 37.41 | 35.46 | 26.58 | 38.56 |
| Noun & Verb | Phrase | 26.87 | 21.37 | 27.51 | 29.67 | 23.67 | 29.92 |
| | Phrase-Morph | 33.19 | 24.28 | 3.03 | 35.49 | 27.58 | 42.73 |

[14] http://nlp.stanford.edu/software/tagger.shtml

The translation systems were evaluated by BLEU score (Papineni et al., 2002). Also, as the reduction in number of unknowns in the translation output indicates better handling of data sparsity, we counted the number of OOV words in the translation outputs. Table 9 shows the BLEU evaluation scores of the translation systems for both Phrase and factor based models and Table 10 shows the OOV reduction numbers statistics.

Table 10: Counts of total OOVs present before morphology injection and the % OOV reduction after Morph Injection

| Morph Problem | Model | # OOV | | | OOV reduction (%) | | |
|---|---|---|---|---|---|---|---|
| | | En - Hi | En - Mr | En-Ml | En - Hi | En - Mr | En-Ml |
| Noun | Fact | 3,030 | 2,399 | 2,706 | 57.39 | 57.08 | 58.02 |
| | Fact-Morph | 1,739 | 1,369 | 1,489 | | | |
| Verb | Fact | 3,041 | 2,772 | 2,894 | 67.78 | 61.14 | 61.42 |
| | Fact-Morph | 980 | 1,695 | 1,534 | | | |
| Noun & Verb | Fact | 3,393 | 4,137 | 4,124 | 55.02 | 39.48 | 55.00 |
| | Fact-Morph | 1,867 | 2,963 | 2,345 | | | |
| Noun & Verb | Phrase | 1,013 | 2,572 | 2,312 | 25.67 | 22.32 | 21.98 |
| | Phrase-Morph | 753 | 1,998 | 1,854 | | | |

Table 11: Counts of total OOVs translated after morphology injection and the matches with the reference used for BLEU evaluation

| Morph Problem | En - Hi | | En - Mr | | En-Ml | |
|---|---|---|---|---|---|---|
| | # OOV translated | # Ref. Matches | # OOV translated | # Ref. Matches | # OOV translated | # Ref. Matches |
| Noun (Fact) | 1291 | 558 | 1030 | 248 | 1217 | 523 |
| Verb (Fact) | 2061 | 971 | 1077 | 253 | 1360 | 642 |
| Noun & Verb (Fact) | 1526 | 687 | 1174 | 284 | 1779 | 613 |
| Noun & Verb (Phrase) | 260 | 71 | 574 | 116 | 458 | 123 |

### 9.2 Subjective Evaluation

As BLEU evaluation with only single reference is not a true measure of evaluating our method, we also performed human evaluation. We found out that Fact-Morph/Phrase-Morph systems really have better outputs compared to Fact/Phrase systems, in terms of both, adequacy and fluency.

Table 12. Subjective evaluation scheme for Adequacy [Ramanathan et al., 2009]

| Level | Interpretation |
|-------|----------------|
| 5 | All meaning is conveyed |
| 4 | Most of the meaning is conveyed |
| 3 | Much of the meaning is conveyed |
| 2 | Little meaning is conveyed |
| 1 | None of the meaning is conveyed |

Table 13. Subjective evaluation scheme for Fluency [Ramanathan et al., 2009]

| Level | Interpretation |
|-------|----------------|
| 5 | Flawless Hindi, with no grammatical errors whatsoever |
| 4 | Good Hindi, with a few minor errors in morphology |
| 3 | Non-native Hindi, with possibly a few minor grammatical errors |
| 2 | Disfluent Hindi, with most phrases correct, but ungrammatical overall |
| 1 | Incomprehensible |

Table 14. Subjective evaluation of the translation systems with and without word-form dictionary

| Morph Problem | Model | Adequacy | | | Fluency | | |
|---------------|-------|----------|-------|-------|---------|-------|-------|
| | | En - Hi | En - Mr | En-Ml | En - Hi | En - Mr | En-Ml |
| Noun | Fact | 34 % | 28 % | 35% | 36 % | 31 % | 35% |
| | Fact-Morph | 56 % | 48 % | 58% | 65.04% | 57.52% | 64.32% |
| Verb | Fact | 38.48 % | 30% | 37.43 % | 48% | 40% | 53% |
| | Fact-Morph | 58.87% | 49.32% | 54.89 % | 72% | 60.78% | 71.23% |
| Noun & Verb | Fact | 39.32% | 34.56% | 38.67 % | 45.05% | 42.01% | 46.02% |
| | Fact-Morph | 49.87% | 45.45% | 51% | 60.04% | 53.32% | 61.34% |
| Noun & Verb | Phrase | 32.38% | 26.34% | 33.87 % | 34.98% | 30.76% | 36.12% |
| | Phrase-Morph | 40.96% | 38.86% | 42.56 % | 58.43% | 55.87% | 64.12% |

For evaluation, randomly chosen 50 translation outputs from each system were manually given adequacy and fluency scores. The scores were given on the scale of 1 to 5 going from worst to best, respectively. Table 14 shows average scores for each system. We observe up to 34.36% improvement in adequacy and up to 44.05% improvement in fluency for the English to Hindi systems and up to 41.67% improvement in adequacy and up to 45.72% improvement in fluency for the English to Marathi systems. Table 12 and Table 13 show the evaluation schemes used [ Ramanathan et al., 2009]. Table 12 shows average adequacy and fluency scores for each system.

From the automatic evaluation scores, it is very evident that Fact-Morph/Phrase-Morph outperforms Fact/Phrase while solving any morphology problem in both Hindi and Marathi. But, improvements in En-Mr systems are low. This is due to the small size of word-form dictionaries that are used for injection. % reduction in OOV shows that, morphology injection is more effective with factored models than with the phrase-based model. Also, improvements shown by BLEU are less compared to % reduction in OOV.

### 9.3 Why BLEU improvement is low?

One possible reason is ambiguity in lexical choice. Word-form dictionary may have word forms of multiple Hindi or Marathi root words for a single parallel English root word. Hence, many times the translation of the English word may not match the reference used for BLEU evaluation, even though it may be very similar in the meaning. Table 11 shows the number of OOVs that are actually translated after morphology injection and number of translated OOVs that match with the reference. We see that matches with the reference are very less compared to the actual number of OOVs translated. Thus, BLEU score cannot truly reflect the usefulness of morphology injection.

From the subjective evaluation scores, we found out that Fact-Morph/Phrase-Morph systems really have better outputs compared to Fact/Phrase systems, in terms of both, adequacy and fluency. We observe up to 34.36% improvement in adequacy and up to 44.05% improvement in fluency for the English to Hindi systems and up to 41.67% improvement in adequacy and up to 45.72% improvement in fluency for the English to Marathi systems.

### 9.4 Why Phrase-based models perform badly?

Factored models showed improvement after morphology injection. But, the performance of phrase-based models degraded. The possible reason may be because in latter case, we are just injecting morphological forms into the corpus without providing any extra information about when to use them.

For example, phrase-based model trained with evidence of only boys-लड़के {ladake}, when augmented with boys-लड़कों {ladakon}, has equal probability to translate boys to लड़के {ladake} or लड़कों {ladakon}. But, factored model trained with the evidence of boys | boy | direct- लड़के {ladake} | लड़का (ladakaa) | ए(e) when augmented with boys | boy | oblique- लड़कों{ladakon} | लड़का{ladakaa} | ओं{on}, can correctly translate boys to लड़के{ladake} or लड़कों{ladakon} based on direct and oblique case.

For example, noun boys} in English can translate to लड़के {ladake} or लड़कों {ladakon} in Hindi. Suppose, we train a phrase-based model with the training data having evidence of only boys-लड़के {ladake}. We also train a factored model as described in Section 4.1 on the same data but with case as an extra factor. Hence, factored training corpus will have evidence of only boys | boy | direct- लड़के {ladake} | लड़का (ladakaa) | ए(e). Now, we inject a word-form boys- लड़कों {ladakon} and

boys|boy|oblique- लड़कों {ladakon}|लड़का {ladakaa}|ओं {on} in the training corpus of phrase-based and factored model, respectively. Then, phrase-based model has equal probability to translate boys to लड़के {ladake} or लड़कों {ladakon}. This ambiguity may lead to incorrect choice of word while translating. On the other hand, factored model knows when to use which form correctly based on direct and oblique case.

### Test cases:

We also performed a qualitative evaluation. We present some examples in Table 15 with detailed explanation of phenomena with case study.

### Table 15: TEST Cases with examples

| Examples | Test Sentences | Explanation of Phenomena |
|---|---|---|
| **Example 1:** | *There is a crowd of traders of the world at the <u>auction center</u>.* | In this case, Fact02 correctly translated auction to नीलाम {neelam}. |
| **Fact:** | वहाँ के <u>auction मध्य</u> में दुनिया के व्यापारियों की भीड़ लगी रहती है ।<br><br>*{vahan ke auction madhya mein vyapariyon ki bhiid lagii rahatii hai.}*<br><br>*{ there auction center in traders crowd is there }* | Also, note that, as Fact01 could not translate *auction*, the next word, *center* is incorrectly translated to मध्य {madhu} {middle}. The correct translation is केंद्र |
| **Fact-Morph:** | वहाँ के नीलाम केंद्र में दुनिया के व्यापारियों की भीड़ लगी रहती है ।<br><br>*{vahan ke niilam kendra mein vyapariyon ki bhiid lagii rahatii hai.}*<br><br>*{there in nilam centre world traders  crowd is there}* | {kendr} {center}. Thus, we also see improvements in the correct lexical choice for the words in local context of the nouns. |
| **Example 2** | <u>*Eyelids*</u> *are a thin fold of skin that cover and protect the eye.* | Again in this case, eyelids and fold are not translated by Fact01, but Fact02 correctly translates them to पलकें {palkem} and गुना {guna}, respectively. |
| **Fact:** | <u>eyelids</u> त्वचा की पतली <u>fold</u> हैं कि और आँखों की रक्षा करते हैं<br><br>*{eyelids tvachaa kii patalii fold hai ki aur aankhon kii rakshaa kartein hain.}*<br><br>*{eyelids skin thin fold and eyes are being protected}* | |
| **Fact-Morph** | <u>पलकें</u> त्वचा की पतली <u>गुना</u> हैं कि और आँखों की रक्षा करते हैं<br><br>*{palaken tvachaa kii patalii gunaa hai ki aur aankhon kii rakshaa kartein hain.}*<br><br>*{palkem skin thin guna and eyes are being protected}* | |

## 10. GENERALIZED SOLUTION

In Section 5, we studied the sparseness problem and its solution in context of solving the noun and verb morphology for English as a source language and Hindi as a target language. But, can the process to generate all factor combinations be generalized for other morphologically richer languages on the target side? We have investigated a generalized solution to this problem. We can use technique for new target language X if:

- We identify the factor set, say S, that affects the inflections of words in language X and can extract them from English sentence

- We know which inflection the target word will have for a particular factor combination of factors in S on source side

- We have a joiner tool in language X to generate the surface word from the root word and suffix

## 11. CONCLUSION AND FUTURE WORK

SMT approaches suffer due to data sparsity while translating into a morphologically rich language. We solve this problem by enriching the original data with the missing morphological forms of words. Morphology injection performs very well and improves the translation quality. We observe huge reduction in number of OOVs and improvement in adequacy and fluency of the translation outputs. We observe up to 34.36% improvement in adequacy and up to 44.05% improvement in fluency for the English to Hindi systems and up to 41.67% improvement in adequacy and up to 45.72% improvement in fluency for the English to Marathi systems. This method is more effective when used with factored models than the phrase-based models. Though the approach of solving data sparsity seems simple, the morphology generation may be painful for target languages which are morphologically too complex. A possible future work is to generalize the approach of morphology generation and verify the effectiveness of morphology injection on morphologically complex languages.

## ACKNOWLEDGMENTS

The authors would like to thank Department of Science and Technology, Govt. of India for the funding under Women Scientist Scheme- WOS-A with the project code- SR/WOS-A/ET-1075/2014.